\def\year{2022}\relax
\documentclass[letterpaper]{article} 
\usepackage{aaai22}  
\usepackage{times}  
\usepackage{helvet}  
\usepackage{courier}  
\usepackage[hyphens]{url}  
\usepackage{graphicx} 
\usepackage{natbib}  
\usepackage{caption} 
\DeclareCaptionStyle{ruled}{labelfont=normalfont,labelsep=colon,strut=off} 
\usepackage{algorithm}
\usepackage{algorithmic}

\usepackage{newfloat}
\usepackage{listings}
\floatstyle{ruled}
\newfloat{listing}{tb}{lst}{}
\floatname{listing}{Listing}
\usepackage{amsmath,amsthm,amscd,amssymb,amsbsy,mathtools}
\usepackage{cancel}
\usepackage{enumitem}
\usepackage{array}
\usepackage{booktabs}
\usepackage{multirow}
\usepackage[caption=false,subrefformat=parens]{subfig}
\usepackage{bbding}
\usepackage{ulem}
\title{Towards End-to-End Image Compression and Analysis with Transformers}
\author{
    Yuanchao Bai$^{1,2}$\footnote{Equal contribution. $\dagger$Corresponding author.}, Xu Yang$^{1*}$, Xianming Liu$^{1,2\dagger}$, Junjun Jiang$^{1,2}$, \\Yaowei Wang$^{2}$, Xiangyang Ji$^{3}$, Wen Gao$^{2,4}$ \\
}
\affiliations{
    $^{1}$Harbin Institute of Technology, $^{2}$Peng Cheng Laboratory, $^{3}$Tsinghua University, $^{4}$Peking University\\
    yuanchao.bai@gmail.com, 20s103274@stu.hit.edu.cn, \{csxm,jiangjunjun\}@hit.edu.cn\\ wangyw@pcl.ac.cn, xyji@tsinghua.edu.cn, wgao@pku.edu.cn
}
\def\x{\mathbf x}
\def\y{\mathbf y}
\def\z{\mathbf z}
\def\h{\mathbf h}
\def\p{\mathbf p}

\def\hx{{\hat{\mathbf x}}}
\def\hy{{\hat{\mathbf y}}}
\def\hz{{\hat{\mathbf z}}}

\def\tz{{\tilde{\mathbf z}}}
\def\tc{{\tilde{\mathbf c}}}

\def\bbE{\mathbb E}
\def\bbR{\mathbb R}

\DeclareMathOperator*{\const}{const}
\DeclareMathOperator*{\FFN}{FFN}
\DeclareMathOperator*{\MSA}{MSA}
\DeclareMathOperator*{\LN}{LN}
\DeclareMathOperator*{\Softmax}{Softmax}

\newcolumntype{C}[1]{>{\centering}p{#1}}

\graphicspath{{./figures/}}
\begin{document}

\maketitle

\begin{abstract}
    We propose an end-to-end image compression and analysis model with Transformers, targeting to the cloud-based image classification application.
    Instead of placing an existing Transformer-based image classification model directly after an image codec, we aim to redesign the Vision Transformer (ViT) model to perform image classification from the compressed features and facilitate image compression with the long-term information from the Transformer.
    Specifically, we first replace the patchify stem (\emph{i.e.}, image splitting and embedding) of the ViT model with a lightweight image encoder modelled by a convolutional neural network.
    The compressed features generated by the image encoder are injected convolutional inductive bias and are fed to the Transformer for image classification bypassing image reconstruction.
    Meanwhile, we propose a feature aggregation module to fuse the compressed features with the selected intermediate features of the Transformer, and feed the aggregated features to a deconvolutional neural network for image reconstruction.
    The aggregated features can obtain the long-term information from the self-attention mechanism of the Transformer and improve the compression performance.
    The rate-distortion-accuracy optimization problem is finally solved by a two-step training strategy.
    Experimental results demonstrate the effectiveness of the proposed model in both the image compression and the classification tasks.

\end{abstract}

\section{Introduction}
Vision Transformer (ViT) \cite{vit2021iclr} and its variations \cite{deit2021icml,wu2021cvt,ceit2021,chen2021visformer,swin2021}, inherited from Transformer architecture \cite{vaswani2017attention} in natural language processing (NLP), have recently demonstrated outstanding performance on a board range of image analysis tasks, such as image classification \cite{vit2021iclr}, segmentation \cite{setr2021cvpr} and object detection \cite{yolo2021vit}.
With the self-attention mechanism, these models are capable of capturing long-range dependencies in the image data, but inevitably result in high computational cost.
In practice, Transformer-based models are usually deployed in the cloud-based paradigm and executed remotely.
For example, massive image data is acquired by the frontend devices, such as mobile phones or surveillance cameras, and transmitted to the cloud (\emph{i.e.}, data center) for further analysis, sharing and storage.
Image compression serves as a fundamental infrastructure for data communication between the frontend and the cloud.

In the traditional paradigm of cloud-based applications, image compression is considered independent of image analysis, and adopts lossy image compression standards designed for \textit{human vision}, such as JPEG \cite{wallace1992jpeg}. In particular, the raw images are first transformed to the frequency domain with Discrete Cosine Transform (DCT). The frequency coefficients are then quantized to discard high frequencies that are less sensitive to human eyes. The quantized coefficients are encoded to bitstreams with entropy encoding and are transmitted to the cloud. On the cloud side, the quantized coefficients are recovered from the received bitstreams, which are then inversely transformed to reconstruct images. The reconstruction distortions are minimized with respect to Peak Signal-to-Noise Ratio (PSNR). However, if the reconstructed images optimized by PSNR are fed into the downstream image analysis tasks, which are tailored to \textit{machine vision} instead, the corresponding results may be inaccurate, because the principle of machine vision is different from human vision \cite{haohan2020cvpr}.
Besides, the traditional image codecs are comprised of hand-crafted modules with complex dependencies. It is difficult to optimize the sophisticated compression frameworks together with subsequent machine analysis tasks.

Recently, learning-based image compression emerges as an active research area in computer vision community.
A number of learning-based image codecs, such as
\cite{Toderici2016iclr,theis2017iclr,li2018cvpr,Balle2018variational,minnen2018nips,cheng2020cvpr,Ma2020pami,hyy2021pami},
have achieved comparable or even better perceptual performance than traditional image codecs for human vision.
Besides, by replacing the hand-crafted modules with deep neural networks (DNNs), learning-based image compression can be integrated with high-level tasks and end-to-end optimized for machine vision \cite{Torfason2018TowardsIU,Chamain2021EndtoEndOI,Le2021ImageCF}.
However, compared with image compression for human vision, image compression for machine vision is still in its infancy, because it is challenging to achieve the best of both worlds for low-level and high-level tasks.

In this paper, we propose a novel paradigm that is friendly for both human vision and machine vision, which integrates learning-based image compression with Transformer-based image analysis. The derived end-to-end image compression and analysis model leads to the synergy effect of these two tasks.
Instead of placing an existing Transformer-based image classification model directly after an image codec, we redesign the ViT model to perform image classification from the compressed features \cite{Torfason2018TowardsIU} and facilitate image compression with the long-term information from the Transformer.
Specifically, we replace the the \textit{patchify stem} (\emph{i.e.}, image splitting and embedding) of the ViT model with a lightweight image encoder modelled by a convolutional neural network (CNN).
The compressed features generated by the image encoder are injected convolutional inductive bias and are more expressive than the features extracted by the patchify stem from the decoded images.
When transmitted to the cloud, the compressed features are fed to the Transformer for image classification bypassing image reconstruction.
We further propose a feature aggregation module to fuse the compressed features with the selected intermediate features of the Transformer, and feed the aggregated features to a deconvolutional neural network for image reconstruction.
The aggregated features obtain the long-term information from the Transformer and effectively improve the compression performance.
We interpret the corresponding \textit{rate-distortion-accuracy} optimization problem based on variational auto-encoder (VAE) \cite{theis2017iclr,Balle2018variational} and information bottleneck (IB) \cite{tishby2000information,vib2017iclr}, and finally solve it with a two-step training strategy.

The main contributions are summarized as follows:
\begin{itemize}
    \item We propose an end-to-end image compression and analysis model, which performs image classification from the compressed features. We interpret the rate-distortion-accuracy optimization problem based on VAE and IB.
    \item We design the network by integrating learning-based image compression with ViT-based image analysis, which leads to the synergy between the two tasks.
    \item In terms of rate-distortion, the proposed model achieves PSNR performance close to BPG \cite{bpg}. In terms of rate-accuracy, the proposed model outperforms ResNet50 \cite{he2016deep}, DeiT-S \cite{deit2021icml} and Swin-T \cite{swin2021} classification from the decoded images, while significantly reduces the computational cost under equivalent number of parameters.
\end{itemize}

\section{Related Work}
\subsubsection{Image Compression for Machine Vision.}
With the fast progress of artificial intelligence, an increasing amount of visual data is now not only viewed by humans but also analyzed by machines.
Recently, image/video compression for machine vision has drawn significant interests in the computer vision community \cite{vcm2020tip}.

In order to optimize image compression with analysis, \cite{quantization2020eccv,Luo2021TheRT} and \cite{quannet2019icme} proposed to optimize the quantization of the traditional codecs JPEG and JPEG2000 to improve the performance of the following image classification. However, since the frameworks of traditional codecs are different from fully optimizable DNN and only the quantization is involved in the optimization, the improvement is limited. In contrast, learning-based image compression is more suitable to be jointly optimized with DNN-based image analysis. The related works can be divided into two categories: 1) \textit{RGB inference}, such as \cite{Chamain2021EndtoEndOI} and \cite{Le2021ImageCF}, performs image analysis from RGB reconstructed images by placing image analysis methods directly after existing image codecs.
2) \textit{Compressed inference}, such as \cite{Torfason2018TowardsIU}, performs image analysis directly from the compressed features bypassing image reconstruction.

In this paper, we propose an end-to-end image compression and analysis model with Transformers, inspired by \cite{Torfason2018TowardsIU}. Beyond \cite{Torfason2018TowardsIU}, we interpret the rate-distortion-accuracy optimization problem based on VAE and IB, and design the Transformer-based model leading to the synergy between the two tasks.

\subsubsection{Transformers in Computer Vision.}
Nowadays, Transformers have shown their potential to be a viable alternative to CNNs in computer vision tasks.
However, the ViT model \cite{vit2021iclr} without any human-defined inductive bias suffers from over-fitting when the training data is limited, and thus needs sophisticated data augmentation schemes \cite{deit2021icml}. In order to improve the performance and the robustness of Transformers, several works \cite{wu2021cvt,ceit2021,chen2021visformer} incorporated CNNs into Transformers.

In this paper, we propose to replace the patchify stem of the ViT model with a CNN-based image encoder, which can enable image analysis from the compressed features and effectively improve the performance of image classification. The concurrent work \cite{earlyconv2021} also observes that early convolutions in Transformers can increase the optimization stability and improve the Top-1 accuracy. Our experimental results are consistent with the observation of \cite{earlyconv2021}.

\section{Proposed Method}
\subsection{Problem Formulation}
\begin{figure}[!t]
\centering
\includegraphics[width=0.9\linewidth]{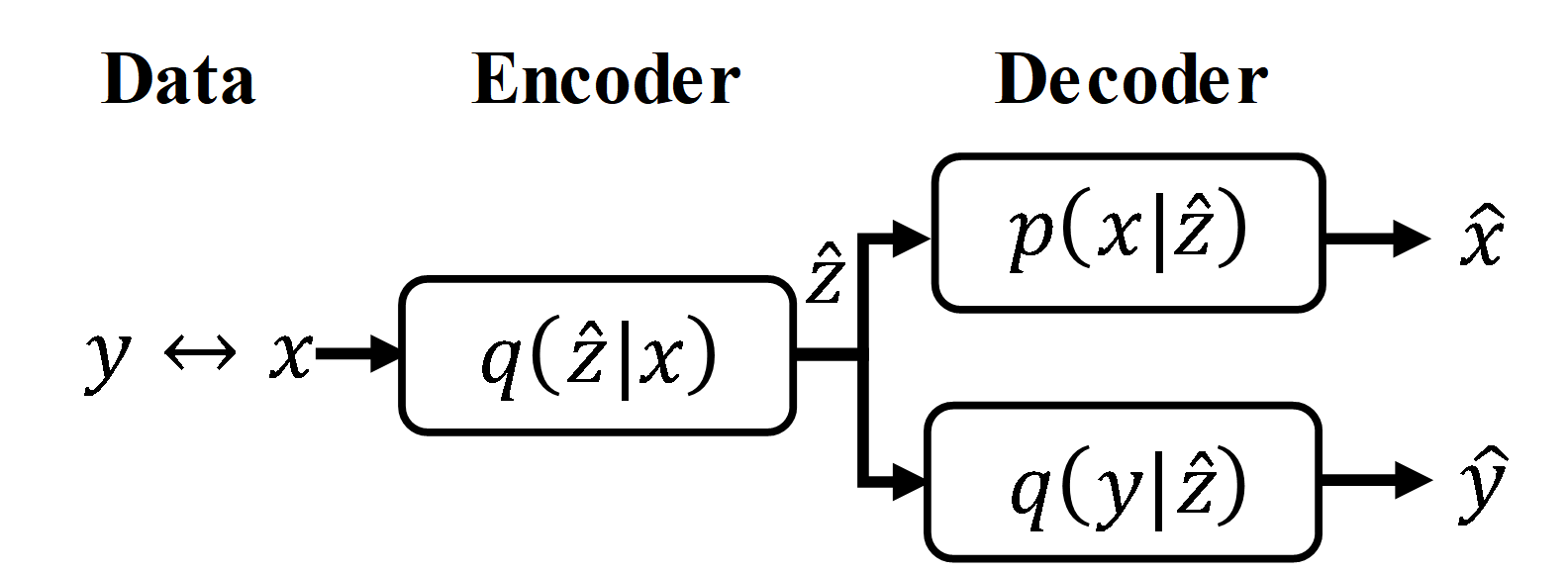}
\caption{The theoretical framework of the proposed end-to-end image compression and analysis model.}
\label{fig:framework}
\end{figure}

We aim to perform image analysis from the compressed features.
Given a raw image $\x$ and its label $\y$, our goal is to learn a compressed representation $\hz$ that facilitates both image decoding (reconstruction) and analysis, as sketched in Fig.\;\ref{fig:framework}.
Since the compressed representation $\hz$ is extracted from the image $\x$ while not accessing the label $\y$, we assume that $\x$, $\y$, $\hz$ form a Markov chain $\y\leftrightarrow\x\leftrightarrow\hz$, leading to $p(\hz|\x,\y)=p(\hz|\x)$.

\subsubsection{Image Compression.}
We first formulate the lossy image compression model without taking image analysis into consideration. Following the standard framework of variational auto-encoder based image compression \cite{theis2017iclr,Balle2018variational}, the latent representation $\z$ is transformed from the raw image $\x$ by an encoder and is quantized to the discrete-valued $\hz$. Then, $\hz$ is losslessly compressed with entropy encoding techniques \cite{arithmetic_coding,duda2009asymmetric} to form a bitstream. On the decoder side, $\hz$ is recovered from the bitstream and inversely transformed to a reconstructed image $\hx$.
To optimize the performance of the compression model, it can be approximated by the minimization of the expectation of Kullback-Leibler (KL) divergence between the intractable true posterior $p(\hz|\x)$ and a parametric inference model $q(\hz|\x)$ over the data distribution $p(\x)$ \cite{Balle2018variational}:
\begin{align}
    \bbE_{p(\x)}D_{kl}[q(\hz|\x)||&p(\hz|\x)]=\bbE_{p(\x)}\bbE_{q(\hz|\x)}[\cancelto{0}{\log q(\hz|\x)} \notag \\
    &-\log p(\x|\hz)-\log p(\hz)]+ \const
    \label{eq:lossy_compression}
\end{align}
where $D_{kl}[\cdot||\cdot]$ denotes KL divergence.
Because the transform from $\x$ to $\z$ is deterministic and the quantization of $\z$ is relaxed by adding noise from uniform distribution $\mathcal{U}(-\frac12,\frac12)$, we have $q(\hz|\x)=\prod_{i} \mathcal{U}(z_{i}-\frac12,z_{i}+\frac12)$ and thus the first term $\log q(\hz|\x)=0$. The second term of \eqref{eq:lossy_compression} is interpreted as the expected distortion between $\x$ and $\hx$, and the third term is interpreted as the cost of encoding $\hz$, leading to the rate-distortion trade-off \cite{shannon1948mathematical}.

\subsubsection{Image Analysis.}
We then turn to consider image analysis. We propose to maximize the mutual information $I(\hz, \y)$ between the compressed representation $\hz$ and the label $\y$, inspired by information bottleneck \cite{tishby2000information,vib2017iclr}.
The mutual information $I(\hz, \y)$ is the reduction in the uncertainty of $\y$ due to the knowledge of $\hz$:
\begin{align}
    I(\hz,\y)&=H(\y)-H(\y|\hz) \notag \\
    &=H(\y)+\sum_{\y, \hz}p(\y, \hz)\log p(\y|\hz)
\end{align}
where $H(\cdot)$ denotes the entropy. Because the true posterior $p(\y|\hz)$ is also intractable, we propose a variational approximation $q(\y|\hz)$, which is the decoder for image analysis apart from the decoder for image reconstruction. Since $D_{kl}[p(\y|\hz)||q(\y|\hz)]\ge0$, we have $\sum_{\y} p(\y|\hz)\log p(\y|\hz) \ge \sum_{\y} p(\y|\hz)\log q(\y|\hz)$ and thus
\begin{equation}
    I(\hz,\y) \ge H(\y)+\sum_{\y, \hz}p(\y, \hz)\log q(\y|\hz)
    \label{eq:I_zy_lb1}
\end{equation}
Because the entropy $H(\y)$ is independent of $\hz$, we can maximize $\sum_{\y, \hz}p(\y, \hz)\log q(\y|\hz)$ as a proxy for $I(\hz,\y)$.
Based on the Markov chain assumption, we replace $p(\y, \hz)$ with $\sum_{\x}p(\x,\y,\hz)=\sum_{\x}p(\x,\y)p(\hz|\x)$, and can rewrite $\sum_{\y, \hz}p(\y, \hz)\log q(\y|\hz)$ as
\begin{equation}
    \bbE_{p(\x,\y)}\bbE_{p(\hz|\x)}\log q(\y|\hz)
    \label{eq:I_zy_lb2}
\end{equation}
With $q(\y|\hz)$, we can generate the estimated label $\hy$ from $\hz$.

\subsubsection{Joint Optimization.}
With \eqref{eq:lossy_compression} and \eqref{eq:I_zy_lb2}, we further formulate the joint optimization of both image compression and analysis. Since $p(\hz|\x)$ in \eqref{eq:I_zy_lb2} is intractable, we share the inference model $q(\hz|\x)$ in \eqref{eq:lossy_compression} as the approximation, and minimize the approximated negative \eqref{eq:I_zy_lb2} together with \eqref{eq:lossy_compression}\footnote{$\bbE_{p(\x)}D_{kl}[q(\hz|\x)||p(\hz|\x)]=\bbE_{p(\x,\y)}D_{kl}[q(\hz|\x)||p(\hz|\x)]$}:
\begin{equation}
    \bbE_{p(\x,\y)}\{-\alpha\bbE_{q(\hz|\x)}\log q(\y|\hz)+ D_{kl}[q(\hz|\x)||p(\hz|\x)]\}
    \label{eq:joint_op1}
\end{equation}
where $\alpha$ is a trade-off parameter. Suppose that $p(\x|\hz)$ is given by $\mathcal{N}(\x|\hx,(2\beta)^{-1}\mathbf{1})$, we can finally rewrite \eqref{eq:joint_op1} to the objective function:
\begin{equation}
    \bbE_{p(\x,\y)}\bbE_{q(\hz|\x)}[-\alpha\log q(\y|\hz)+\beta\|\x-\hx\|^2_2-\log p(\hz)]
    \label{eq:joint_op2}
\end{equation}
The first term of \eqref{eq:joint_op2} weighted by $\alpha$ can be interpreted as the cross-entropy loss for image analysis, such as image classification or segmentation, based on the types of label $\y$. In this work, we choose image classification as the target task. The second term weighted by $\beta$ is the mean square error (MSE) distortion loss. The third term is the rate loss.

In contrast to the image compression models \cite{theis2017iclr,Balle2018variational}, the compressed representation $\hz$ in \eqref{eq:joint_op2} is also optimized for image analysis tasks.
The complexity of $\hz$ in \eqref{eq:joint_op2}
is controlled by minimizing the cost of encoding $\hz$, rather than controlled by minimizing $I(\hz, \x)$ in the information bottleneck models \cite{tishby2000information,vib2017iclr}.

\subsection{Transformer-based Network Architecture}
We realize the theoretical framework in Fig.\;\ref{fig:framework} by proposing an end-to-end image compression and analysis model with Transformers. The proposed model can promote the synergy between the two tasks.

\begin{figure}[!t]
\centering
\includegraphics[width=0.98\linewidth]{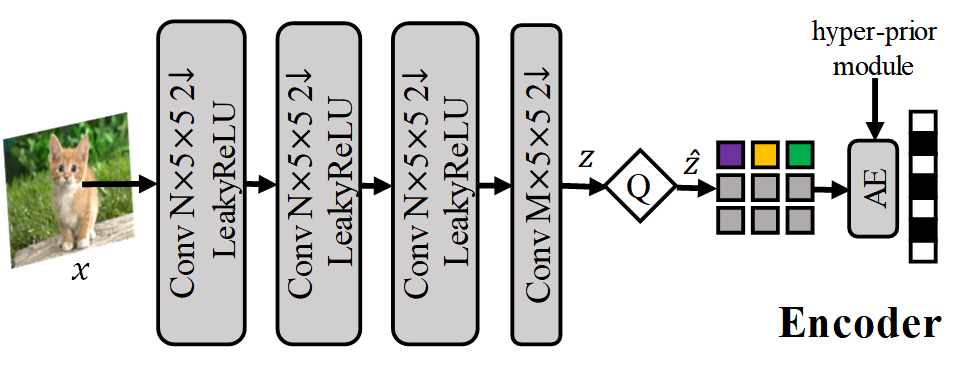}
\caption{The network architecture of the encoder. We set $N=128$ and $M=192$ same as the low-rate setting of \cite{Balle2018variational}. The encoder replaces the patchify stem of ViT and is relatively lightweight, which can be deployed at the frontend. \textbf{Q}: Quantization. \textbf{AE}: Arithmetic Encoder.}
\label{fig:encoder}
\end{figure}

\subsubsection{Encoder.}
The network architecture of the proposed encoder is illustrated in Fig.\;\ref{fig:encoder}. Similar to the setting of \cite{Balle2018variational}, we employ four stride-$2$ $5\times5$ convolutional layers to extract features with gradually reduced spatial resolution from the input image $\x\in\bbR^{H\times W\times3}$.
We use LeakyReLU as the activation function instead of using the Generalized Divisive Normalization (GDN) \cite{gdn2016iclr}, because GDN results in convergence problem when training with the Transformer blocks in the proposed model.

\begin{figure*}[!t]
\centering
\includegraphics[width=0.98\linewidth]{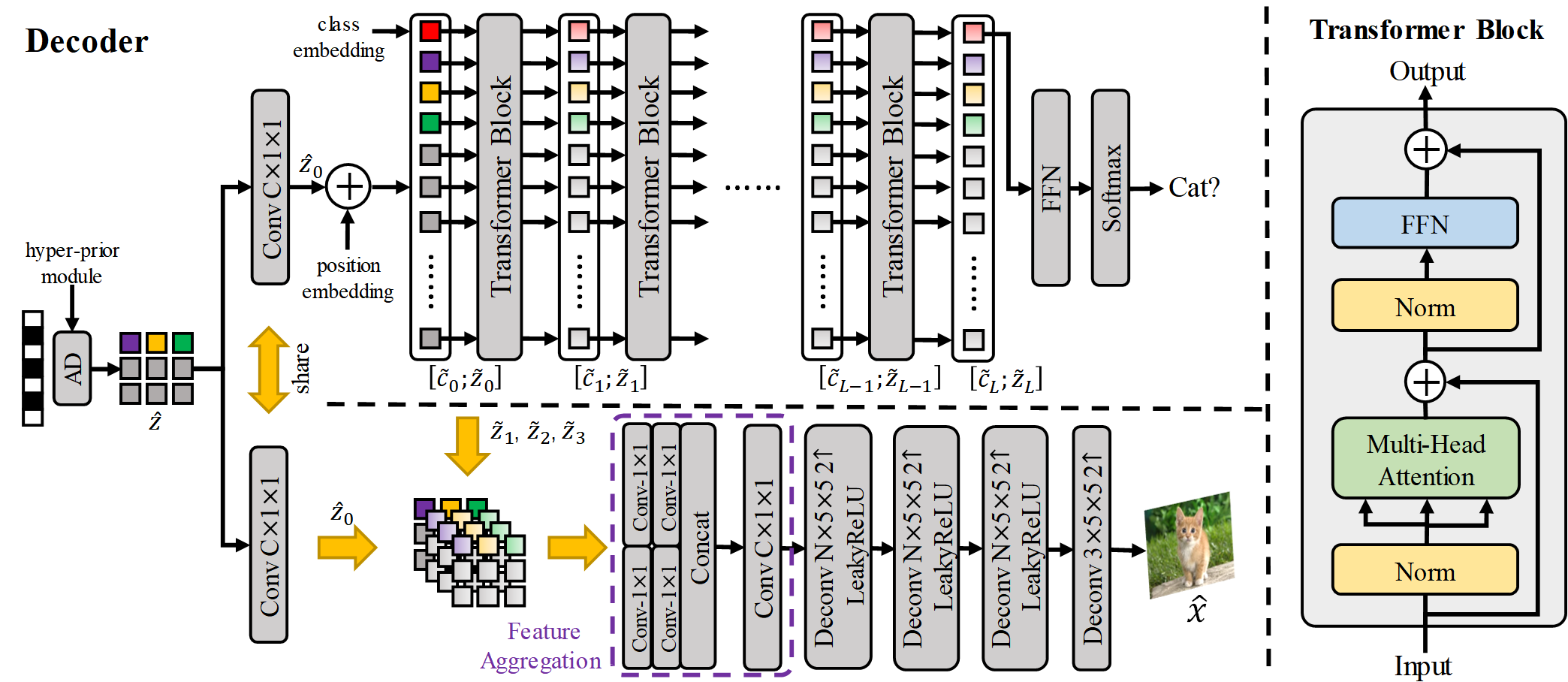}
\caption{The network architecture of the decoder. We set $C=384$, $L=12$ and $N=128$. The number of parameters is equivalent to ResNet50 \cite{he2016deep}. The decoder is deployed on the cloud to classify or reconstruct images from the received bitstreams. \textbf{AD}: Arithmetic Decoder.}
\label{fig:decoder}
\end{figure*}

The resulting feature $\z\in\bbR^{\frac{H}{16}\times \frac{W}{16}\times M}$ is then quantized to the discrete-valued $\hz$. We employ the hyper-prior module \cite{Balle2018variational,minnen2018nips} to estimate $p(\hz|\h_{\hz})$ for the entropy encoding of $\hz$, where $\h_{\hz}$ denotes the hyper-prior of $\hz$. We do not use the serial autoregressive module \cite{minnen2018nips}, because its corresponding decoding time is too long on large-scale image classification datasets.

From the perspective of Transformer architecture design, the proposed encoder can be considered as the replacement of the patchify stem, \emph{i.e.}, a stride-$16$ $16\times16$ convolutional layer, applied to the input image in the ViT model \cite{vit2021iclr}. Several concurrent works \cite{wu2021cvt,ceit2021,chen2021visformer,earlyconv2021} also replace the patchify stem with a stack of convolutional layers, in order to improve the performance of image classification. In the experiments, we observe that the proposed encoder is capable of extracting compressed features suitable for both image decoding reconstruction and classification through joint optimization.
Note that the proposed encoder is relatively lightweight, thus can be deployed at the frontend, such as mobile phones or surveillance cameras.

\subsubsection{Decoder-Classifier.}

The decoder receives the bitstream from the encoder and adopts the shared hyper-prior module $p(\hz|\h_{\hz})$ to recover $\hz$.
For image classification, we directly feed $\hz$ to an inference network, instead of the common subsequent approach---first reconstruct the decoded image $\hx$ and then conduct inference on $\hx$.

Specifically, we adopt the standard Transformer blocks in the ViT model \cite{vit2021iclr} with the number of parameters equivalent to ResNet50 \cite{he2016deep}, as shown in Fig.\;\ref{fig:decoder}.
We expand the channel dimension of $\hz$ to $C$ with a $1\times1$ convolutional layer, and reshape the resulting feature $\hz_0\in\bbR^{\frac{H}{16}\times \frac{W}{16} \times C}$ to a sequence $\hz_0\in\bbR^{\frac{HW}{16^2}\times C}$. To maintain the spatial information of the feature $\hz_0$, we add learnable position embeddings $\p$ to $\hz_0$ leading to $\tz_0=\hz_0+\p$. Following \cite{vit2021iclr}, we prepend a learnable class embedding $\tc_0$, and feed the sequence $[\tc_0; \tz_0]\in\bbR^{(\frac{HW}{16^2}+1)\times C}$ to the Transformer consisting of $L$ Transformer blocks. The architecture of each Transformer block is illustrated in Fig.\;\ref{fig:decoder}. The computation process can be formulated as
\begin{align}
    [\tc'_i;\tz'_i] &= \MSA(\LN([\tc_{i-1};\tz_{i-1}]))+[\tc_{i-1};\tz_{i-1}] \\
    [\tc_i;\tz_i] &= \FFN(\LN([\tc'_i;\tz'_i]))+[\tc'_i;\tz'_i]~~~~ i=1,\ldots, L \notag
\end{align}
where $\MSA(\cdot)$ denotes the multi-head self-attention module, $\FFN(\cdot)$ denotes the feed forward network and $\LN(\cdot)$ denotes the layer normalization \cite{LN2016}, respectively.

With the self-attention mechanism, the class embedding $\tc_i$ interacts with the image feature $\tz_i$, and the final output $\tc_L$ is used to compute $q(\y|\hz)$ for image classification:
\begin{equation}
    q(\y|\hz) = \Softmax(\FFN(\LN(\tc_L)))
\end{equation}
where $\Softmax(\cdot)$ denotes softmax operation. The $\FFN(\cdot)$ is the classifier head mapping the embedding dimension from $C$ to the number of classes.

\subsubsection{Decoder-Reconstructor.}
Reconstructing image $\hx$ directly from $\hz$ (or $\hz_0$) ignores the global spatial correlations among the latent features. Recent image compression works \cite{qian2021iclr,guo2021csvt} demonstrate that leveraging global context information during entropy coding can improve the compression performance.
Transformers naturally capture the global spatial information among the latent features, which can also benefits low-level tasks, such as image processing \cite{ipt2021cvpr} and image generation \cite{Jiang2021TransGANTT}.
Motivated by these works, we aim to extract the intermediate features $\tz_i$'s of the Transformer and incorporate them into image reconstruction.

Specifically, we select $\hz_0$ and $[\tz_1, \tz_2, \tz_3]$, and propose a feature aggregation module to fuse these features, similar to \cite{setr2021cvpr}. Selecting $[\tz_1, \tz_2, \tz_3]$ means that the first three Transformer blocks are also involved in the image reconstruction process.
Since image reconstruction may work independently of image classification, we avoid using $\{\tz_i(i>3)\}$ that involve too many Transformer blocks in the image reconstruction, in order to reduce the computational complexity. The feature aggregation module is illustrated in Fig.\;\ref{fig:decoder}. The computational process can be formulated as
\begin{align}
    \hz''_0&=\mathrm{Conv}_1^0(\hz_0),~~ \tz''_1=\mathrm{Conv}_1^1(\tz_1) \notag\\
    \tz''_2&=\mathrm{Conv}_1^2(\tz_2),~~ \tz''_3=\mathrm{Conv}_1^3(\tz_3) \\
    &\hz_f = \mathrm{Conv}_2([\hz''_0;~\tz''_1;~\tz''_2;~\tz''_3]) \notag
\end{align}
where $\mathrm{Conv}_1^i(\cdot)$ denotes a $1\times 1$ convolutional layer reducing the channel dimension of the input to $\frac{C}{4}$. $\mathrm{Conv}_2(\cdot)$ is another $1\times 1$ convolutional layer with $C$ channels fusing the four concatenated input features. $\hz_f$ is the fused feature.

Finally, we input the fused feature $\hz_f$ to four stride-$2$ $5\times 5$ deconvolutional layers gradually increasing the spatial resolution, leading to the reconstructed RGB image $\hx$.

\subsection{Training Strategy}
We observe that the one-step training strategy, \emph{i.e.}, minimizing \eqref{eq:joint_op2} to train the encoder and decoder from scratch, leads to convergence problem in the experiments. Instead, we employ a two-step training strategy:

\begin{enumerate}[label=\arabic*)]
    \item We pretrain the proposed model without considering the quantization of $\z$ and the hyper-prior module of $\hz$. We remove the rate loss in \eqref{eq:joint_op2} temporarily, and minimize the cross-entropy loss together with the MSE loss. Because the value of the cross-entropy loss is much smaller than that of the MSE loss, we set $\alpha=1$ and $\beta=0.001$ in \eqref{eq:joint_op2} to balance the contributions of the two losses.
    \item We load the pretrained parameters and minimize \eqref{eq:joint_op2} to train the entire network including the quantization of $\z$ and the hyper-prior module of $\hz$. The $\alpha$ and $\beta$ in \eqref{eq:joint_op2} are tuned with fixed $\frac{\alpha}{\beta}$ to achieve different bit rates.
\end{enumerate}

\section{Experiments}
\subsection{Experimental Settings}
\subsubsection{Datasets.}
We perform extensive experiments on the ImageNet dataset \cite{deng2009imagenet} and iNaturalist19 (INat19) dataset \cite{inat19}. ImageNet is well-known image classification dataset containing $1000$ object classes with $1,281,167$ training images and $50,000$ validation images. INat19 is a fine-grained classification dataset containing $1010$ species of plants and animals with $265,213$ training images and $3030$ validation images.

\subsubsection{Pretraining w/o Compression.}
As aforementioned, we pretrain the proposed model without the quantization of $\z$ and the hyper-prior module of $\hz$. We remove the rate loss, and minimize the cross entropy loss together with the MSE loss. We set $\alpha=1$ and $\beta=0.001$, respectively. We set the input size to $224\times224$ and adopt the same data augmentation as DeiT \cite{deit2021icml}, except for the Exponential Moving Average (EMA) \cite{polyak1992acceleration}, which do not enhance the performance of the proposed model. The input images are normalized with ImageNet default mean and standard deviation, and are denormalized during image reconstruction. We observe that random erasing \cite{Zhong2020RandomED}, mixup \cite{Zhang2017mixup} and cutmix \cite{Yun2019CutMixRS} designed for the training of image classification are also compatible with the training of image reconstruction in our experiments.

On the ImageNet dataset, we train the proposed network from scratch. We use AdamW optimizer \cite{adamW2019iclr} for $300$ epochs with minibatches of size $1024$. We set the initial learning rate to $0.001$ and use a cosine decay learning rate scheduler with $5$ epochs warm-up.

On the INat19 dataset, we initialize the network with the pretrained parameters on the ImageNet dataset. The classifier head is adjusted to the class number of INat19. We use AdamW optimizer for $100$ epochs with minibatches of size $512$. We set the initial learning rate to $0.0005$ and use a cosine decay learning rate scheduler with $2$ epochs warm-up.

\subsubsection{Training w/ Compression.}
We load the pretrained parameters on the ImageNet and INat19 datasets, respectively. We recover the quantization of $\z$ and the hyper-prior module of $\hz$. We fix $\frac{\alpha}{\beta}=100$ and set $\alpha\in\{0.1, 0.3, 0.6\}$. We observe that the hyper-prior module of $\hz$ is sensitive to data augmentation, and thus we only employ RandomResizedCropAndInterpolation and RandomHorizontalFlip during training.

On the ImageNet dataset, we load the corresponding pretrained parameters, and use Adam optimizer \cite{kingma2015adam} with a initial learning rate of $0.0001$, following \cite{Balle2018variational}.
We train the proposed network for $300$ epochs with minibatches of size $1024$, and use a cosine decay learning rate scheduler with $5$ epochs warm-up.

On the INat19 dataset, we load the corresponding pretrained parameters, and also use Adam optimizer with a initial learning rate of $0.0001$.
We train the proposed network for $300$ epochs with minibatches of size $512$, and use a cosine decay learning rate scheduler with $2$ epochs warm-up.

\subsection{Experimental Results}
\subsubsection{Pretrained Model.}
Table\;\ref{tb:results_pretrain}(a) reports the experimental results of our pretrained model without compression on ImageNet. We compare with the existing image classification models including CNN-based models, such as ResNet50 \cite{he2016deep} and RegNetY-4G \cite{Radosavovic2020DesigningND}, and Transformer-based models, such as ViT-B \cite{vit2021iclr}, DeiT-S \cite{deit2021icml}, CvT-13 \cite{wu2021cvt}, CeiT-S \cite{ceit2021}, Visformer-S \cite{chen2021visformer}, Swin-T \cite{swin2021} and ViT$_C$-4GF \cite{earlyconv2021}. We select the specific settings of the models with the number of parameters closest to ResNet50.

\begin{table}[!tb]
\small
\centering
\begin{tabular}{p{6em}|C{3.3em}C{3.3em}C{3em}C{3.3em}}
\toprule[1pt]
\multicolumn{5}{c}{\textbf{(a) Results on ImageNet Dataset}} \tabularnewline
  Model & input size & Params (M) & Top-1 (\%)& PSNR (dB) \tabularnewline
\hline
\hline
     ResNet50 & 224  & 25.6 & 75.9 & $-$ \tabularnewline
     RegNetY-4G & 224 & 20.6 & 80.0 & $-$ \tabularnewline
     ViT-B    & 224  & 86.5 & 77.9 & $-$ \tabularnewline
     DeiT-S   & 224  & 22.1 & 79.9 & $-$ \tabularnewline
     CvT-13   & 224  & 20.0 & 81.6 & $-$ \tabularnewline
     CeiT-S   & 224  & 24.2 & 82.0 & $-$ \tabularnewline
     Visformer-S  & 224  & 40.2 & 82.3 & $-$ \tabularnewline
     Swin-T   & 224  & 28.3 & 81.2 & $-$ \tabularnewline
     ViT$_C$-4GF & 224 & 17.8 & 81.4 & $-$ \tabularnewline
\hline
     Ours     & 224  & 25.6 & 81.7 & 31.7 \tabularnewline
\end{tabular}

\begin{tabular}{p{6em}|C{3.3em}C{3.3em}C{3.3em}C{3.3em}}
\toprule[1pt]
\multicolumn{5}{c}{\textbf{(b) Results on INat19 Dataset}} \tabularnewline
  Model & input size & Params (M) & Top-1 (\%)& PSNR (dB) \tabularnewline
\hline
\hline
     ResNet50 & 224  & 25.6 & 71.8 & $-$ \tabularnewline
     DeiT-S   & 224  & 22.1 & 76.2 & $-$ \tabularnewline
     Swin-T   & 224  & 28.3 & 77.9 & $-$ \tabularnewline
\hline
     Ours     & 224  & 25.6 & 78.0 & 31.5 \tabularnewline
\bottomrule[1pt]
\end{tabular}
\caption{Results of our pretrained model without compression on the ImageNet/INat19 datasets, compared with the existing image classification models.}
\label{tb:results_pretrain}
\end{table}

Table\;\ref{tb:results_pretrain}(b) reports the experimental results of our pretrained model on INat19 dataset, compared with ResNet50, DeiT-S and Swin-T.
ResNet50 is finetuned based on \cite{he2016deep}. DeiT-S and Swin-T are finetuned using the same setting as our pretrained model.

From Table\;\ref{tb:results_pretrain}, our pretrained model can achieve comparable or better Top-1 accuracies than the existing image classification models, which demonstrates the efficacy of the image encoder for the Transformer-based image classification.
In terms of image reconstruction evaluated by PSNR, our pretrained model achieves $31.7$ dB and $31.5$ dB, respectively. All these results demonstrate that the pretrained model can provide a satisfactory initialization for the following training with compression.

\subsubsection{Full Model.}
\begin{figure}[!t]
\centering
\subfloat[Rate-distortion and rate-accuracy results on ImageNet.]{
\includegraphics[width=0.99\linewidth]{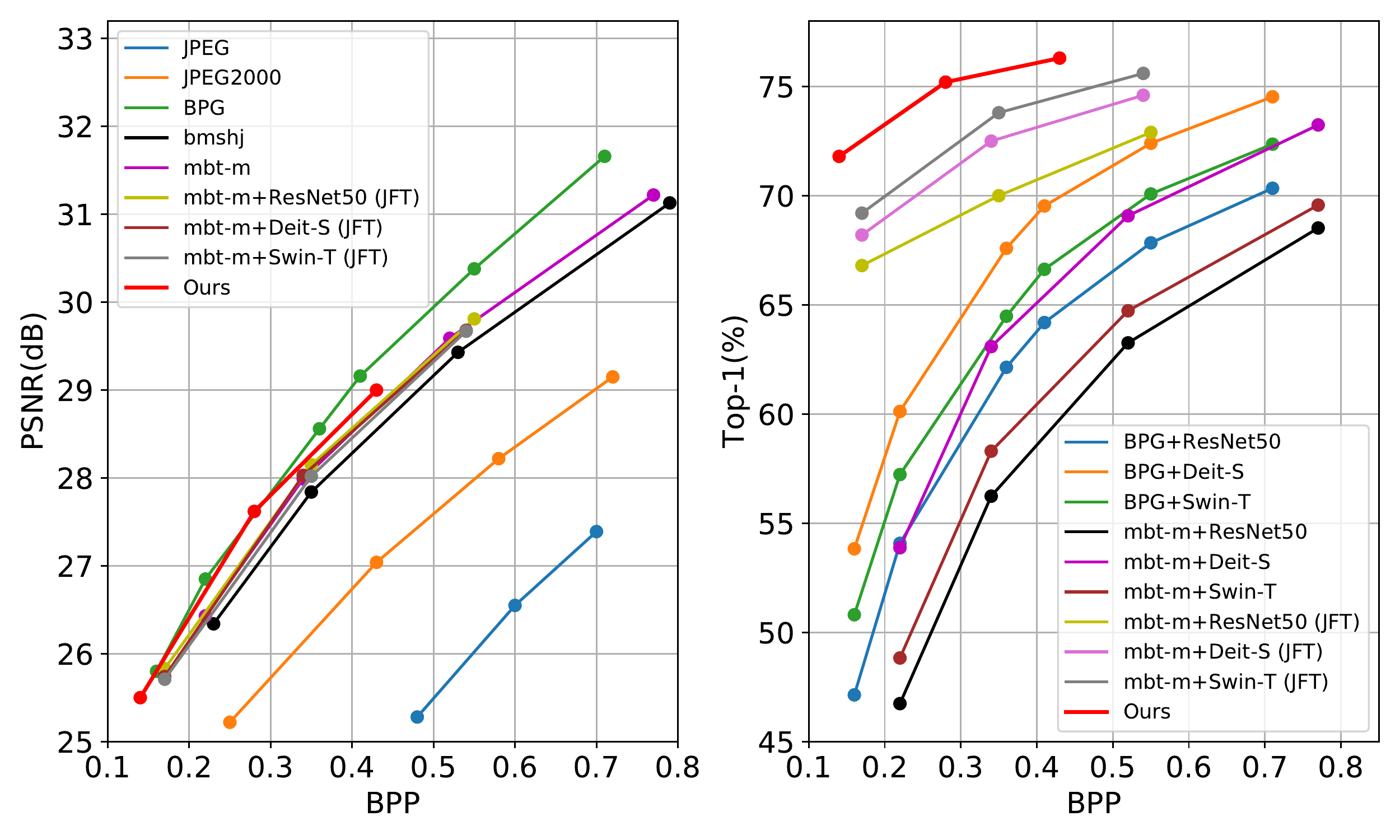}}\\
\subfloat[Rate-distortion and rate-accuracy results on INat19.]{
\includegraphics[width=0.99\linewidth]{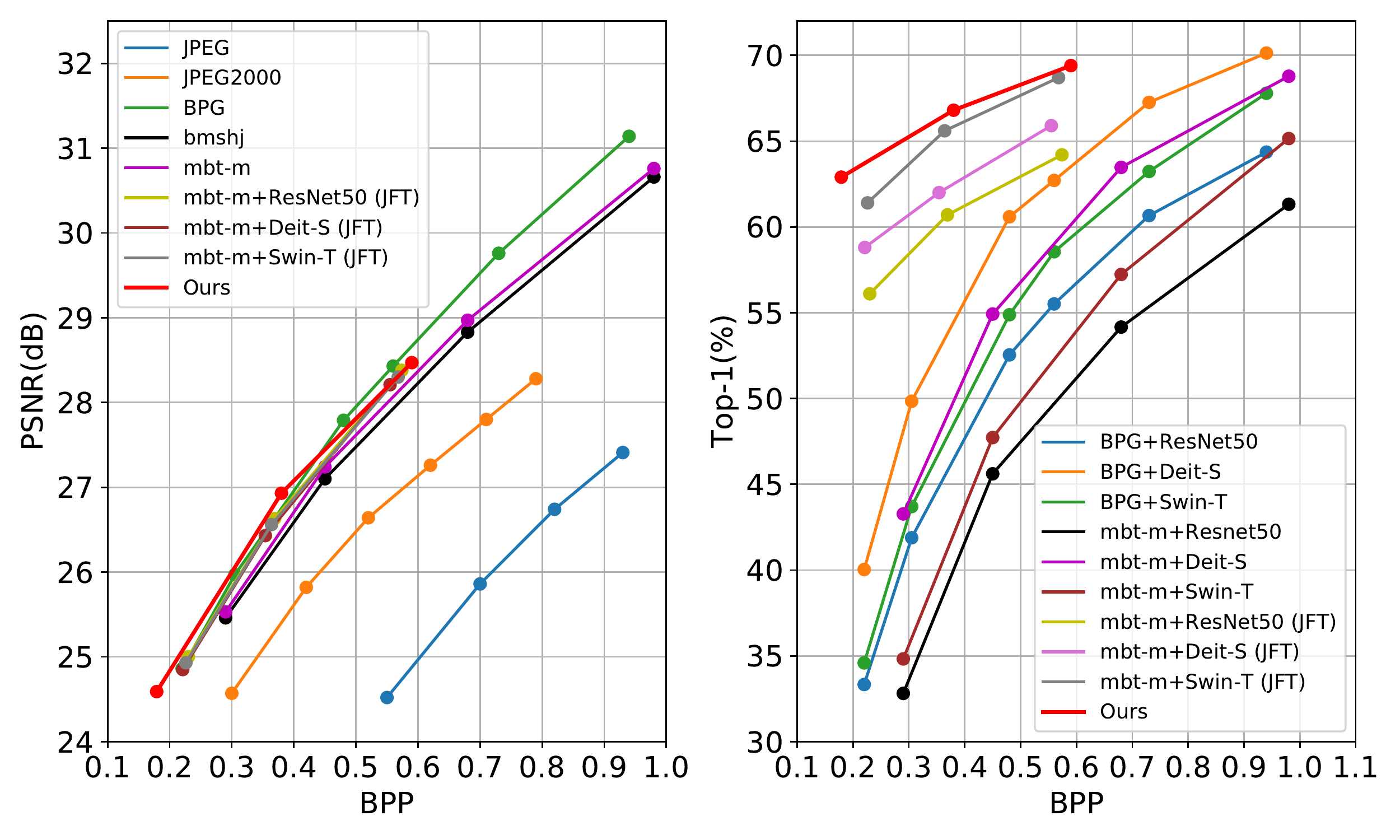}}
\caption{Rate-distortion and rate-accuracy results of the proposed model, compared with the existing image codecs and the image classification methods applied to the reconstructed RGB images. JFT means joint finetune.}
\label{fig:rad_curves}
\end{figure}

In Fig.\;\ref{fig:rad_curves}, we report the rate-distortion and rate-accuracy results of our full model on ImageNet and INat19. We compare with the existing image codecs and the image classification models applied to the decoded images.

We select the traditional image codecs including JPEG \cite{wallace1992jpeg}, JPEG2000 \cite{skodras2001j2k}, BPG \cite{bpg}, and the learning-based image codecs including \textit{bmshj} \cite{Balle2018variational} and \textit{mbt-m} \cite{minnen2018nips}.
The \textit{mbt-m} removes the serial autoregressive module, avoiding long decoding time on the large-scale datasets.
The sophisticated learning-based image codecs with complex entropy models and network architectures, such as \cite{hu2020coarse,cheng2020cvpr,qian2021iclr,guo2021csvt}, are too time-consuming to be evaluated on the large-scale datasets, despite their good compression performance.

We select the decoded images of the best performed traditional and learning-based image codecs in our experiments, \emph{i.e.}, BPG and \textit{mbt-m}, and adopt the image classification models ResNet50, DeiT-S and Swin-T to compute the Top-1 accuracies in comparison with the proposed model.
Moreover, we jointly finetune \textit{mbt-m} together with ResNet50, Deit-S and Swin-T by minimizing \eqref{eq:joint_op2} with $\frac{\alpha}{\beta}=100$, same as the proposed model.

In terms of the rate-distortion performance, the proposed model significantly outperforms the traditional image codecs JPEG and JPEG2000. It is comparable to BPG at relatively low bit-rates but is surpassed by BPG at high bit-rates. The proposed model outperforms the learning-based image codecs \textit{bmshj} and \textit{mbt-m}.
The rate-distortion performance of the jointly finetuned \textit{mbt-m} is similar to the original \textit{mbt-m}. The proposed model also outperforms the jointly finetuned \textit{mbt-m}.

In terms of the rate-accuracy performance, the proposed model outperforms ResNet50, DeiT-S and Swin-T applied to the decoded images of BPG and \textit{mbt-m}, because the image classification models trained on the original datasets are not robust to the decoded images at low bit-rates.
Although Swin-T outperforms DeiT-S on the original datasets (Table.\;\ref{tb:results_pretrain}), DeiT-S outperforms Swin-T on the decoded images at low bit-rates.
After jointly finetuned, Swin-T surpasses DeiT-S on the decoded images at the corresponding bit-rates.
The proposed model also outperforms the jointly finetuned ResNet50, Swin-T and DeiT-S, because the image reconstruction constrained by MSE loss damages the high-frequency information effective for the image classification. The proposed model bypasses the image reconstruction process and directly do inference from the compressed features, which can utilize these high-frequency information.

\begin{figure}[!t]
\centering
\subfloat[JPEG]{
\includegraphics[width=0.32\linewidth]{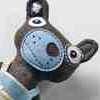}}
\subfloat[JPEG2000]{
\includegraphics[width=0.32\linewidth]{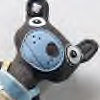}}
\subfloat[BPG]{
\includegraphics[width=0.32\linewidth]{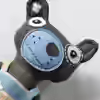}}\\
\subfloat[mbt-m]{
\includegraphics[width=0.32\linewidth]{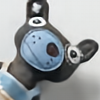}}
\subfloat[mbt-m (JFT)]{
\includegraphics[width=0.32\linewidth]{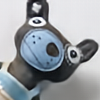}}
\subfloat[Ours]{
\includegraphics[width=0.32\linewidth]{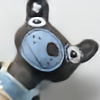}}
\caption{(a) JPEG+DeiT-S, 29.3dB, 0.60bpp, maraca ($\times$). (b) JPEG2000+DeiT-S, 29.17dB, 0.41bpp, can opener ($\times$). (c) BPG+DeiT-S, 29.41dB, 0.18bpp, can opener ($\times$). (d) mbt-m+DeiT-S, 29.79dB, 0.18bpp, can opener ($\times$). (e) mbt-m+Swin-T(JFT), 29.75dB, 0.19bpp, reel ($\times$). (f) Ours, 29.52dB, 0.19bpp, teddy bear ($\checkmark$).}
\label{fig:subjective}
\end{figure}

In Fig.\;\ref{fig:subjective}, we show an illustrative example of the experimental results. Under similar PSNRs, our model achieves comparable or less bit-per-pixel (bpp) and the correct classification result, compared with the other methods.

\subsection{Ablation Studies}
\subsubsection{Rate-Distortion-Accuracy Trade-off.}
\begin{figure}[!t]
\centering
\includegraphics[width=0.99\linewidth]{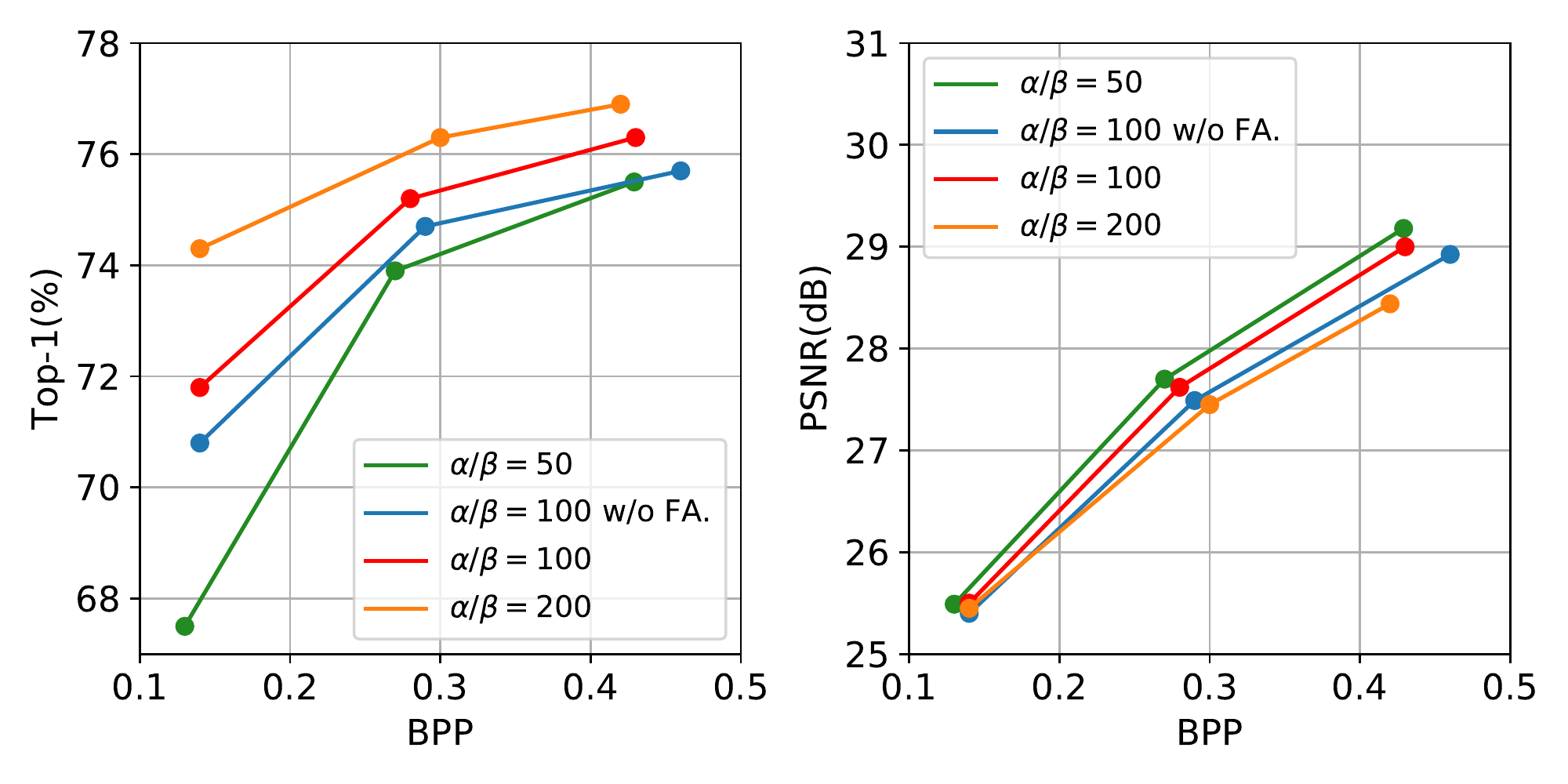}
\caption{Ablation studies of rate-distortion-accuracy trade-off and the feature aggregation (FA) module.}
\label{fig:rad_ablation}
\end{figure}

When the bit-rates of $\hz$ is constrained, minimizing \eqref{eq:joint_op2} with different $\frac{\alpha}{\beta}$ leads to bit allocation between image classification and reconstruction. We empirically set $\frac{\alpha}{\beta}\in\{50, 100, 200\}$ and test the rate-distortion-accuracy trade-off as shown in Fig.\;\ref{fig:rad_ablation}. We can observe that larger $\frac{\alpha}{\beta}$ leads to better Top-1 accuracy but sacrifices PSNR. In contrast, smaller $\frac{\alpha}{\beta}$ leads to better PSNR but lower Top-1 accuracy.

\subsubsection{Feature Aggregation.}
In Fig.\;\ref{fig:rad_ablation}, we compare with the proposed model without the feature aggregation module, \emph{i.e.}, directly feeding $\hz_0$ to the deconvolutional neural network for image reconstruction.
Although the feature aggregation module is only applied to the image reconstruction process, it can benefit both the image classification and reconstruction through rate-distortion-accuracy optimization. The reduced bit-rates from the image reconstruction are potentially allocated to the image classification, leading to the improvement of both tasks.
Although the improvement of Top-1 accuracy is more obvious than PSNR in Fig.\;\ref{fig:rad_ablation}, the decrease of the cross entropy loss $-\alpha\log q(\y|\hz)$ is actually similar to that of the MSE loss $\beta\|\x-\hx\|^2_2$ in \eqref{eq:joint_op2} in the experiments.

\subsubsection{Computational Cost.}
\begin{figure}[!t]
\centering
\includegraphics[width=0.99\linewidth]{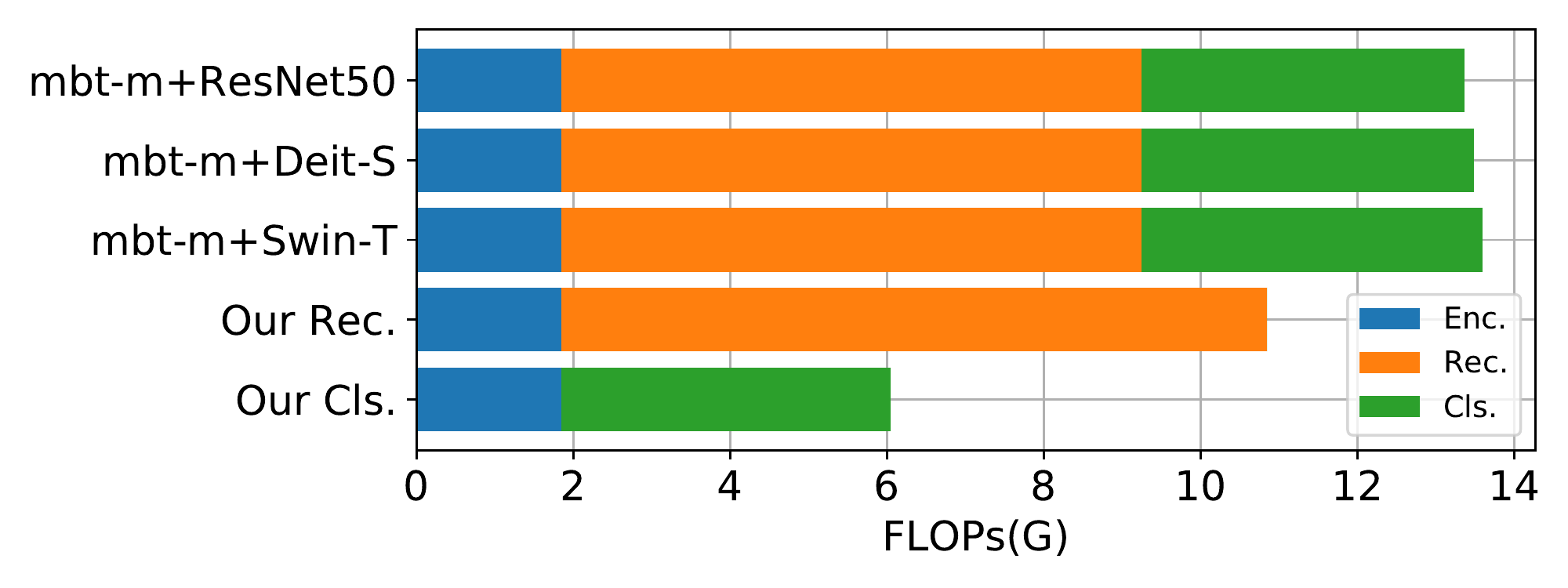}
\caption{Ablation study of the computational cost.}
\label{fig:flops}
\end{figure}

In Fig.\;\ref{fig:flops}, we compare the computational cost of the proposed model with the concatenation of the learning-based image codec \textit{mbt-m} and the image classification methods including ResNet50, DeiT-S and Swin-T.
The architecture of the proposed encoder is similar to the low-rate setting of \textit{mbt-m}, thus their computational costs of image encoding are similar.
In terms of image reconstruction on the decoder side, our image reconstructor needs more FLOPs than \textit{mbt-m} due to the feature aggregation module.
In terms of image classification on the decoder side, our image classifier directly performs inference from the compressed features without the image reconstruction process, and thus needs far less computational cost compared with the inference from reconstructed RGB images.

\section{Conclusion}
In this paper, we learn an end-to-end image compression and analysis model with Transformers, targeting to the cloud-based image classification application.
At the frontend, a CNN-based image encoder extracts compressed features from raw images and transmits them to the cloud.
On the cloud, the compressed features injected convolutional inductive bias are directly fed to the Transformer for image classification bypassing image reconstruction.
Meanwhile, the intermediate features of the Transformer capturing global information are aggregated with the compressed features for image reconstruction.
Experimental results demonstrate the effectiveness of the proposed model in both rate-distortion and rate-accuracy performance.

\section{Appendices}
\subsection{Hyper-prior Module}
\begin{figure}[htb]
\centering
\includegraphics[width=0.9\linewidth]{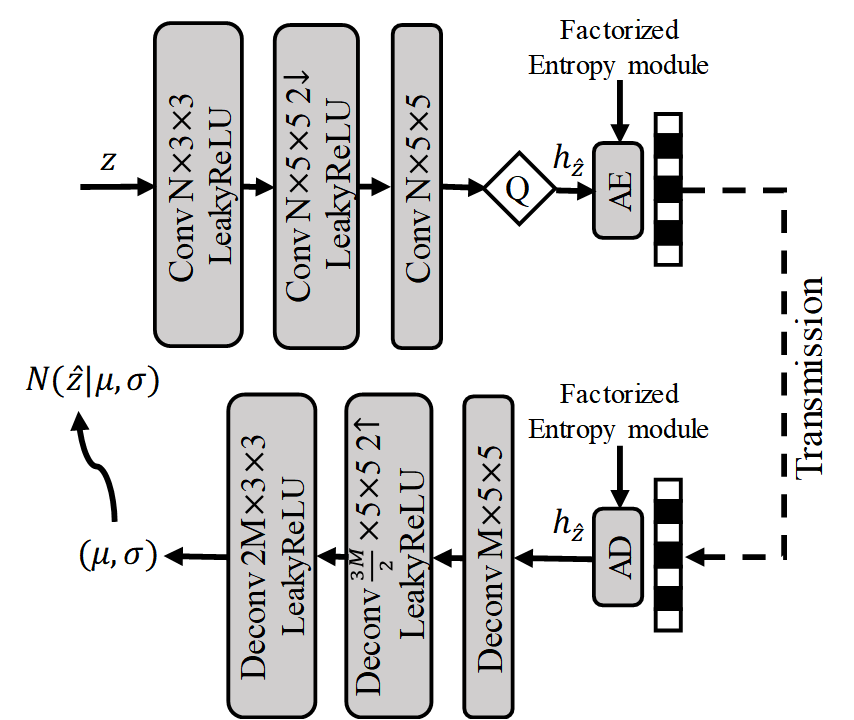}
\caption{The network architecture of the hyper-prior module. \textbf{AE}: Arithmetic Encoding. \textbf{AD}: Arithmetic Decoding.}
\label{fig:hyperprior}
\end{figure}

In the proposed model, we employ the hyper-prior module \cite{Balle2018variational,minnen2018nips} to estimate $p(\hz|\h_{\hz})$ for the entropy encoding of $\hz$, where $\h_{\hz}$ denotes the hyper-prior of $\hz$. Similar to \cite{minnen2018nips}, we assume $p(\hz|\h_{\hz})=\mathcal{N}(\hz|\boldsymbol{\mu}, \boldsymbol{\sigma})$ where the $\boldsymbol{\mu}$ and $\boldsymbol{\sigma}$ are estimated by the neural network, as shown in Fig.\;\ref{fig:hyperprior}. The $\h_{\hz}$ is entropy encoded to the bitstream based on the factorized entropy module of $\h_{\hz}$ and is transmitted to the cloud together with the bitstream of $\hz$.

\subsection{Computing Infrastructure}
The proposed model is trained with eight NVIDIA Tesla V100 GPUs on the ImageNet dataset and is trained with four NVIDIA Tesla V100 GPUs on the INat19 dataset.
The proposed model is evaluated with one NVIDIA RTX3090 GPU on both the datasets.

\subsection{More Ablation Studies}
\subsubsection{Image Reconstruction in Pretraining.}
We report the results of our pretrained model with and without image reconstruction in Table\;\ref{tb:ablation_rec}.
The pretrained model without image reconstruction is a pure image classification model similar to \cite{earlyconv2021}.
Compared with such model, the pretrained model with image reconstruction achieves similar image classification performance.
Meanwhile, the pretrained module with image reconstruction achieves 31.7 dB and 31.5 dB PSNRs on the two datasets, respectively.
These results demonstrate that the pretrained model with image reconstruction can provide a satisfactory initialization for the following training with compression.

\begin{table}[htb]
\centering
\begin{tabular}{p{6em}|C{3em}C{3em}C{3em}C{3em}}
\toprule[1pt]
\multicolumn{5}{c}{\textbf{(a) Results on ImageNet Dataset}} \tabularnewline
  Model & input size & Params (M) & Top-1 (\%)& PSNR (dB) \tabularnewline
\hline
\hline
     Ours w/o rec.     & 224  & 23.3 & 80.7 & $-$ \tabularnewline
     Ours w/ rec.    & 224  & 25.6 & 81.7 & 31.7 \tabularnewline
\end{tabular}

\begin{tabular}{p{6em}|C{3em}C{3em}C{3em}C{3em}}
\toprule[1pt]
\multicolumn{5}{c}{\textbf{(b) Results on INat19 Dataset}} \tabularnewline
  Model & input size & Params (M) & Top-1 (\%)& PSNR (dB) \tabularnewline
\hline
\hline
     Ours w/o rec.     & 224  & 23.3 & 78.4 & $-$ \tabularnewline
     Ours w/ rec.    & 224  & 25.6 & 78.0 & 31.5 \tabularnewline
\bottomrule[1pt]
\end{tabular}

\caption{Effects of the image reconstruction module in the pretrained model.}
\label{tb:ablation_rec}
\end{table}

\subsubsection{Transformer Blocks in Image Reconstruction.}
In Fig.\;\ref{fig:rad_ablation}, we show the efficacy of the feature aggregation module. We further show the efficacy of each Transformer block during image reconstruction in the feature aggregation module.
Based on the learned model, we gradually remove the Transformer blocks by setting the corresponding features to zeros, as shown in Table\;\ref{tb:ablation_tb}. The PSNR decreases with the removal of the Transformer blocks, which demonstrates that each Transformer block plays an important role in the image reconstruction process.

\begin{table}[!t]
\centering
\begin{tabular}{C{4.7em}C{4.7em}C{4.7em}C{4.7em}}
\toprule[1pt]
  $\hz_0$/\sout{$\tz_1$}/\sout{$\tz_2$}/\sout{$\tz_3$} & $\hz_0$/$\tz_1$/\sout{$\tz_2$}/\sout{$\tz_3$} & $\hz_0$/$\tz_1$/$\tz_2$/\sout{$\tz_3$} & $\hz_0$/$\tz_1$/$\tz_2$/$\tz_3$ \tabularnewline
\hline
\hline
     26.49  & 27.31 & 27.42 & 27.62 \tabularnewline
\bottomrule[1pt]
\end{tabular}

\caption{The corresponding PSNRs resulting from removing the Transformer blocks in the feature aggregation module during image reconstruction based on the learned model.}
\label{tb:ablation_tb}

\end{table}

\subsection{More Visualization Results}
\subsubsection{Latent Feature.}
\begin{figure}[!t]
\centering
\includegraphics[width=0.98\linewidth]{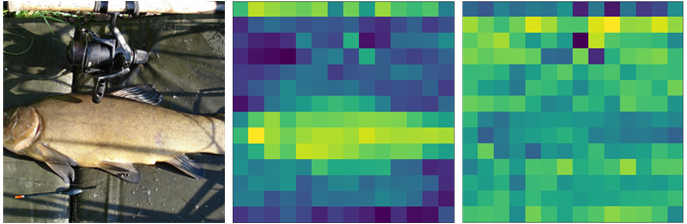}
\caption{Visualization of the latent feature $\hz$.}

\label{fig:feature_map}
\end{figure}

We provide a visual example of the latent feature $\hz$ as shown in Fig.\;\ref{fig:feature_map}.
Fig.\;\ref{fig:feature_map}-\textit{middle} corresponds to a low-frequency channel, which mainly has high responses to the tench.
Differently, Fig.\;\ref{fig:feature_map}-\textit{right} corresponds to a high-frequency channel.

\subsubsection{Reconstruction and Classification Results.}
\begin{figure*}[!t]
\centering
\subfloat[JPEG]{
\includegraphics[width=0.32\linewidth]{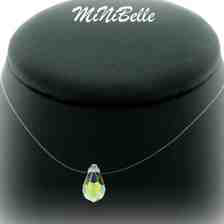}}
\subfloat[JPEG2000]{
\includegraphics[width=0.32\linewidth]{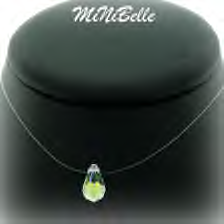}}
\subfloat[BPG]{
\includegraphics[width=0.32\linewidth]{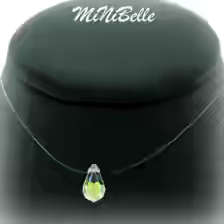}}\\
\subfloat[mbt-m]{
\includegraphics[width=0.32\linewidth]{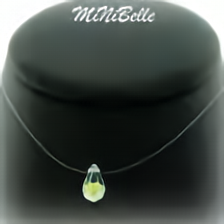}}
\subfloat[mbt-m (JFT)]{
\includegraphics[width=0.32\linewidth]{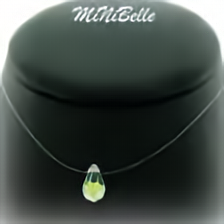}}
\subfloat[Ours]{
\includegraphics[width=0.32\linewidth]{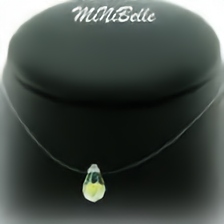}}
\caption{(a) JPEG+DeiT-S, 32.78dB, 0.46bpp, face powder ($\times$). (b) JPEG2000+DeiT-S, 32.34dB, 0.26bpp, face powder ($\times$). (c) BPG+DeiT-S, 33.13dB, 0.11bpp, face powder ($\times$). (d) mbt-m+DeiT-S, 32.98dB, 0.16bpp, soap dispenser ($\times$). (e) mbt-m+Swin-T(JFT), 32.99dB, 0.16bpp, puck/hockey ($\times$). (f) Ours, 32.87dB, 0.13bpp, necklace ($\checkmark$).}
\label{fig:subjective1}
\end{figure*}

\begin{figure*}[!t]
\centering
\subfloat[JPEG]{
\includegraphics[width=0.32\linewidth]{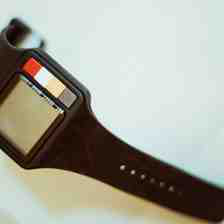}}
\subfloat[JPEG2000]{
\includegraphics[width=0.32\linewidth]{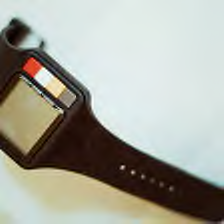}}
\subfloat[BPG]{
\includegraphics[width=0.32\linewidth]{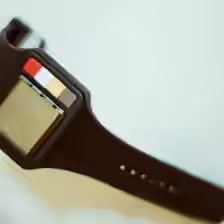}}\\
\subfloat[mbt-m]{
\includegraphics[width=0.32\linewidth]{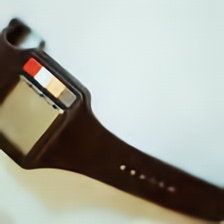}}
\subfloat[mbt-m (JFT)]{
\includegraphics[width=0.32\linewidth]{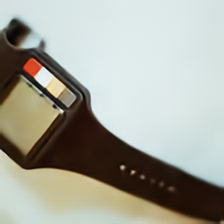}}
\subfloat[Ours]{
\includegraphics[width=0.32\linewidth]{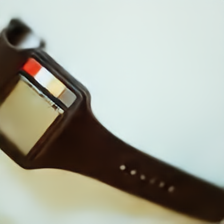}}
\caption{(a) JPEG+DeiT-S, 31.71dB, 0.48bpp, switch/electric switch ($\times$). (b) JPEG2000+DeiT-S, 32.35dB, 0.32bpp, buckle ($\times$). (c) BPG+DeiT-S, 32.55dB, 0.09bpp, hand-held computer ($\times$). (d) mbt-m+DeiT-S, 33.15dB, 0.13bpp, buckle ($\times$). (e) mbt-m+Swin-T(JFT), 33.28dB, 0.14bpp, clog/geta/patten/sabot ($\times$). (f) Ours, 32.99dB, 0.13bpp, digital watch ($\checkmark$).}
\label{fig:subjective2}
\end{figure*}

\begin{figure*}[!t]
\centering
\subfloat[JPEG]{
\includegraphics[width=0.32\linewidth]{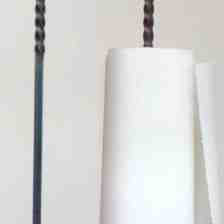}}
\subfloat[JPEG2000]{
\includegraphics[width=0.32\linewidth]{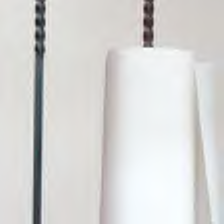}}
\subfloat[BPG]{
\includegraphics[width=0.32\linewidth]{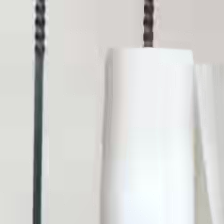}}\\
\subfloat[mbt-m]{
\includegraphics[width=0.32\linewidth]{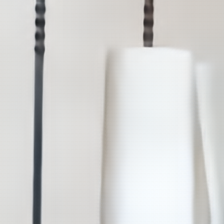}}
\subfloat[mbt-m (JFT)]{
\includegraphics[width=0.32\linewidth]{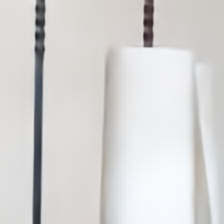}}
\subfloat[Ours]{
\includegraphics[width=0.32\linewidth]{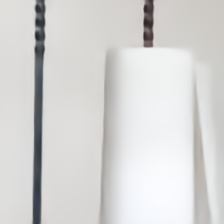}}
\caption{(a) JPEG+DeiT-S, 36.32dB, 0.37bpp, modem ($\times$). (b) JPEG2000+DeiT-S, 36.33dB, 0.17bpp, table lamp ($\times$). (c) BPG+DeiT-S, 36.3dB, 0.05bpp, table lamp ($\times$). (d) mbt-m+DeiT-S, 37.34dB, 0.07bpp, shovel ($\times$). (e) mbt-m+Swin-T(JFT), 37.2dB, 0.07bpp, lampshade ($\times$). (f) Ours, 37.19dB, 0.07bpp, paper towel ($\checkmark$).}
\label{fig:subjective3}
\end{figure*}

\begin{figure*}[!t]
\centering
\subfloat[JPEG]{
\includegraphics[width=0.32\linewidth]{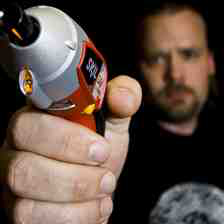}}
\subfloat[JPEG2000]{
\includegraphics[width=0.32\linewidth]{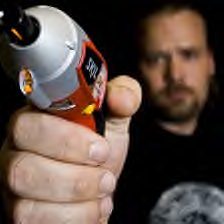}}
\subfloat[BPG]{
\includegraphics[width=0.32\linewidth]{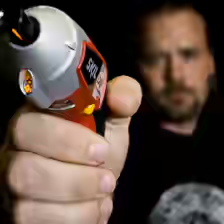}}\\
\subfloat[mbt-m]{
\includegraphics[width=0.32\linewidth]{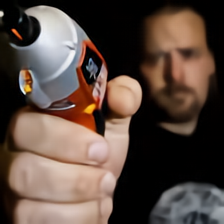}}
\subfloat[mbt-m (JFT)]{
\includegraphics[width=0.32\linewidth]{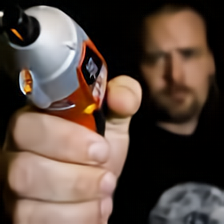}}
\subfloat[Ours]{
\includegraphics[width=0.32\linewidth]{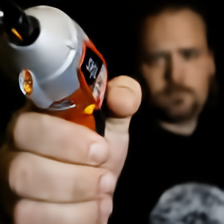}}
\caption{(a) JPEG+DeiT-S, 29.83dB, 0.66bpp, can opener ($\times$). (b) JPEG2000+DeiT-S, 30.43dB, 0.48bpp, can opener ($\times$). (c) BPG+DeiT-S, 30.06dB, 0.21bpp, can opener ($\times$). (d) mbt-m+DeiT-S, 30.29dB, 0.22bpp, lighter/light/igniter/ignitor ($\times$). (e) mbt-m+Swin-T(JFT), 30.34dB, 0.23bpp, stop watch ($\times$). (f) Ours, 30.13dB, 0.21bpp, power drill ($\checkmark$).}
\label{fig:subjective5}
\end{figure*}

\begin{figure*}[!t]
\centering
\subfloat[JPEG]{
\includegraphics[width=0.32\linewidth]{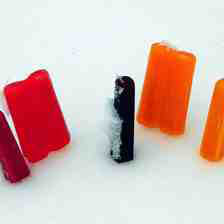}}
\subfloat[JPEG2000]{
\includegraphics[width=0.32\linewidth]{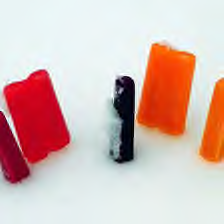}}
\subfloat[BPG]{
\includegraphics[width=0.32\linewidth]{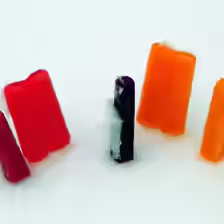}}\\
\subfloat[mbt-m]{
\includegraphics[width=0.32\linewidth]{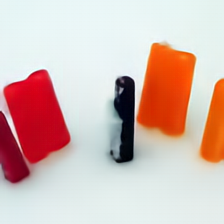}}
\subfloat[mbt-m (JFT)]{
\includegraphics[width=0.32\linewidth]{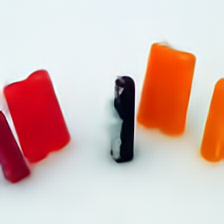}}
\subfloat[Ours]{
\includegraphics[width=0.32\linewidth]{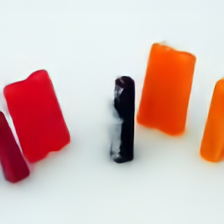}}
\caption{(a) JPEG+DeiT-S, 30.29dB, 0.47bpp, whistle ($\times$). (b) JPEG2000+DeiT-S, 31.76dB, 0.35bpp, whistle ($\times$). (c) BPG+DeiT-S, 32.08dB, 0.1bpp, whistle ($\times$). (d) mbt-m+DeiT-S, 32.73dB, 0.13bpp, whistle ($\times$). (e) mbt-m+Swin-T(JFT), 32.79dB, 0.13bpp, whistle ($\times$). (f) Ours, 32.55dB, 0.13bpp, ice lolly/lolly/lollipop/popsicle ($\checkmark$).}
\label{fig:subjective6}
\end{figure*}

In Fig.\;\ref{fig:subjective1}-\ref{fig:subjective6}, we show more visualization examples of the experimental results. Under similar PSNRs, the visual results of the proposed model are better than the traditional image codecs (\emph{i.e.}, JPEG, JPEG2000, BPG), and comparable to learning-based mbt-m \cite{minnen2018nips} and mbt-m (JFT). JFT denotes joint finetune. Meanwhile, the proposed model achieves comparable or less bit-per-pixel (bpp) and the correct classification, compared with the other methods.

\section{Acknowledgement}
This work was supported by National Key Research and Development Project under Grant 2019YFE0109600, National Natural Science Foundation of China under Grants 61922027, 61827804, 6207115, 61971165 and U20B2052, PCNL Key Project under Grant PCL2021A07, China Postdoctoral Science Foundation under Grant 2020M682826.

\bibliography{aaai22}

\end{document}